\documentclass[letterpaper, 10 pt, conference]{ieeeconf}  
\IEEEoverridecommandlockouts
\usepackage[nolist]{acronym}
\usepackage{algorithm}
\usepackage{algpseudocode}
\usepackage{amsmath}
\usepackage{amssymb}
\usepackage{bm}
\usepackage{cite}
\usepackage{graphicx}
\usepackage{siunitx}
\usepackage{tikz}

\newacro{ai}[AI]{Artificial Intelligence}
\newacro{auroc}[AUROC]{Area Under the Receiver Operating Curve}
\newacro{ml}[ML]{Machine Learning}
\newacro{lg}[LG]{Level Ground}
\newacro{sa}[SA]{Stair Ascent}
\newacro{sd}[SD]{Stair Descent}
\newacro{rf}[RF]{Random Forest}
\newacro{rnn}[RNN]{Recurrent Neural Network}
\newacro{ssl}[SSL]{Self-Supervised Learning}
\newacro{tcn}[TCN]{Temporal Convolutional Network}
\newacro{tf}[TF]{Transformer}
\newacro{rnn}[RNN]{Recurrent Neural Network}
\newacro{imu}[IMU]{Inertial Measurement Unit}

\title{\LARGE \bf
User-Tailored Learning to Forecast Walking Modes for Exosuits
}

\author{Gabriele Abbate$^{1}$, Enrica Tricomi$^{2}$, Nathalie Gierden$^{2}$, Alessandro Giusti$^{1}$, Lorenzo Masia$^{2}$, Antonio Paolillo$^{1}$
\thanks{This work was supported by Innosuisse - Swiss Innovation Agency, through the VRHEM project (100.533 IP-ICT), by the Swiss National Science Foundation (grant number 213074), by the Multidimension AI Project from the Carl-Zeiss Foundation (P2022-08-010),  and by the Istituto nazionale per l'assicurazione contro gli infortuni sul lavoro (INAIL) under grant agreement PR23-RR-P1 FeatherEXO.
}%
\thanks{$^{1}$Dalle Molle Institute for Artificial Intelligence (IDSIA), USI-SUPSI, Lugano, Switzerland {\tt\small name.surname@idsia.ch}}%
\thanks{$^{2}$Munich Institute for Robotics and Machine Intelligence (MIRMI), Department of Computer Engineering, School of Computation, Information and Technology, Technical University of Munich (TUM), Munich, Germany.}
}

\begin{document}

\maketitle
\thispagestyle{empty}
\pagestyle{empty}

\begin{abstract}
Assistive robotic devices, like soft lower-limb exoskeletons or exosuits, are widely spreading with the promise of helping people in everyday life. To make such systems adaptive to the variety of users wearing them, it is desirable to endow exosuits with advanced perception systems. However, exosuits have little sensory equipment because they need to be light and easy to wear. This paper presents a perception module based on machine learning that aims at estimating 3 walking modes (i.e., ascending or descending stairs and walking on level ground) of users wearing an exosuit. We tackle this perception problem using only inertial data from two sensors. Our approach provides an estimate for both future and past timesteps that supports control and enables a self-labeling procedure for online model adaptation.
Indeed, we show that our estimate can label data acquired online and refine the model for new users.
A thorough analysis carried out on real-life datasets shows the effectiveness of our user-tailored perception module. Finally, we integrate our system with the exosuit in a closed-loop controller, validating its performance in an online single-subject experiment.
\end{abstract}

\section{Introduction}\label{sec:intro}

Soft exosuits are rapidly expanding as rehabilitation and assistive devices for their lightness, ease of use, and effectiveness in improving walking endurance~\cite{xiloyannis2021, kim2022, ishmael2021, hood2022}.
To fulfill their goal, exosuits must have advanced perception skills, e.g. to enable adaptive behavior
and cope with the variety of users wearing them~\cite{slade2022}.
Furthermore, it is desirable to provide exosuits' controllers with predictive capabilities to enable real-time adjustments to users' intention and improve responsiveness~\cite{jayakumar2023} and user experience~\cite{lu2024}.

\begin{figure}
\centering
\frame{\includegraphics[trim={0 6cm 9cm 10.5cm},clip,height=0.54\linewidth]{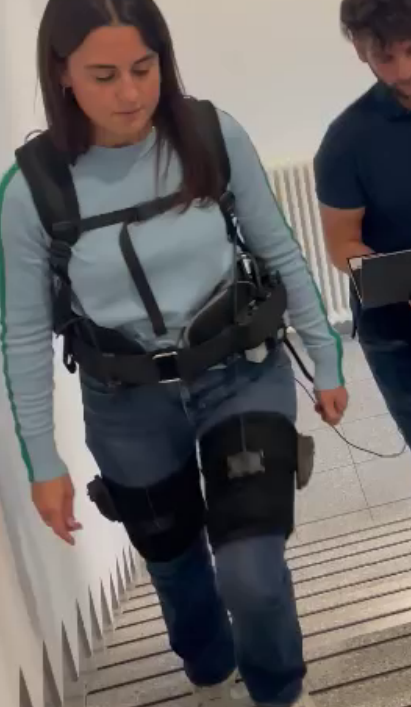}}
\hfill
\frame{\includegraphics[trim={0cm 1cm 4.5cm 5.0cm},clip,height=0.54\linewidth]{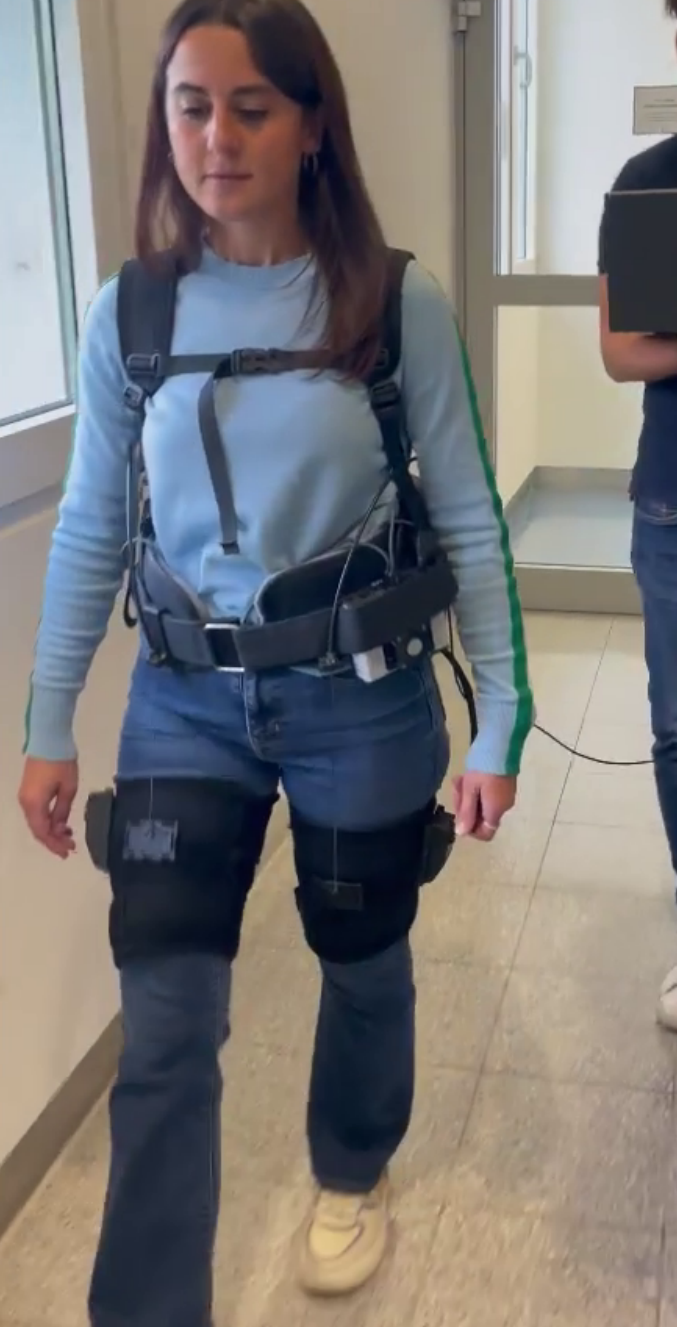}}
\hfill
\frame{\includegraphics[trim={3cm 1cm 3cm 10cm},clip,height=0.54\linewidth]{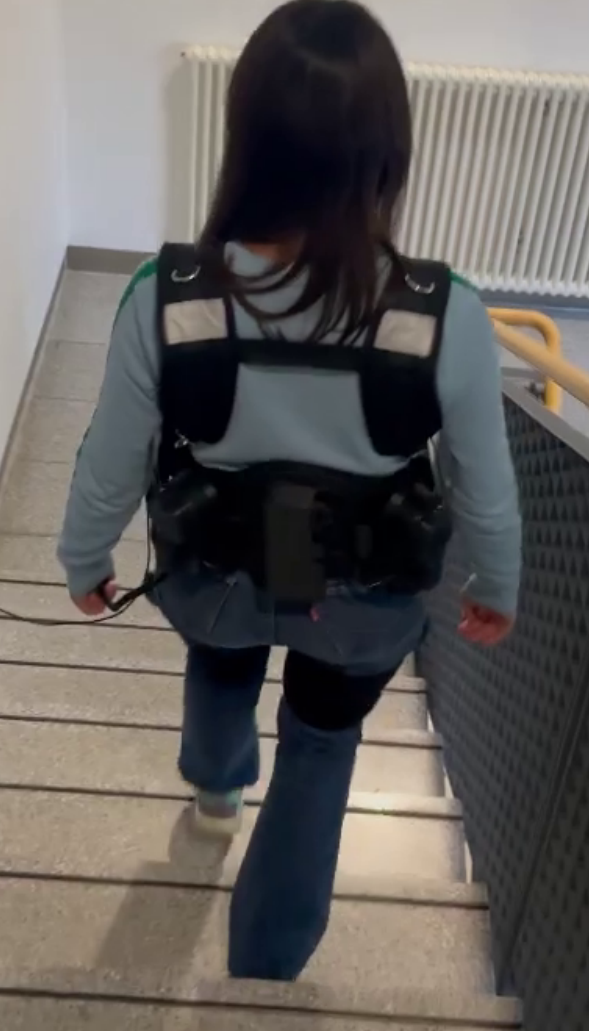}}
\caption{We equip an easy-to-wear and light exosuit with a user-tailored perception model estimating the current and intended walking gait (ascending or descending stairs and walking on level ground) using only IMU readings.}
\label{fig:presentation}
\end{figure}

However, light exosuits have inherently scarce sensory equipment, which makes the development of perception modules a challenge.
%
To this end, \ac{ai} and \ac{ml} offer suitable tools to extract meaningful information from scarce, noisy, and raw sensor data.
\ac{ml} is commonly used for robot perception, for example, in the domain of human-robot interaction~\cite{Abbate:ras:2024}, or rehabilitation robotics~\cite{Yip:science:2023}.
In assistive robotics, \ac{ml} is used to detect walking modes (e.g., walking on level ground or ascending/descending stairs) to adapt the control of underactuated exosuits to the user's motion.
However, many works rely on additional hardware which can either be complex to integrate into a lightweight device (e.g., load cell and pressure insole sensors~\cite{park2014design, liu2020real}) or have technical limitations (like the approaches using cameras~\cite{tricomi2023environment} that suffer from occluded views and difficult light conditions).
In this regard, \ac{imu} sensors are a sensible option, since they are lightweight and can be effectively deployed to detect changes in walking patterns~\cite{yuan2014fuzzy}.
Even if their measurements are prone to angular drift~\cite{siviy2023opportunities}, \ac{ml} methods can be used on \ac{imu} orientation data to classify the walking modes, as presented in a previous work~\cite{zhang:tmec:2024}.
In this paper, we provide an \ac{ml} method to classify 3 walking modes of users wearing an exosuit using only \ac{imu} measurements, see Fig.~\ref{fig:presentation}.
With respect to Zhang et al.~\cite{zhang:tmec:2024}, we propose to: ($i$) reduce the number of sensors from $3$ to $2$, to make the system even lighter; ($ii$) instead of a \ac{rf} we rely on a \ac{tcn} architecture for its capability of handling raw temporal data, without the need for handcrafting relevant features;
($iii$), we extend the model to produce estimates not only at the current time but within a window extending a few seconds in the future and the past.

Another crucial challenge for assistive wearable devices is their adaptability to new users, which is key to their effectiveness~\cite{Poggensee:science:2021}.  
This capability is difficult to implement with traditional model-based control methods since they rely on predefined dynamics that can vary significantly from one individual to another~\cite{Luo2021}. At the same time, \ac{ml}-based approaches require a substantial amount of data to ensure robust performance across different users~\cite{Rose2021, Kang2022}.
To achieve this capability, our pre-trained perception model can be fine-tuned on a new user with data acquired \emph{during} the operation of the device, without any explicit supervision.  In particular, pseudo-labels are derived for a given timestep \emph{in hindsight}, exploiting data collected in the following few seconds. These labels are saved in a user-specific dataset and used to adapt the model to the user automatically.
%
%
This approach is a form of \emph{Self-Supervised Robot Learning}, 
which allows a robot to collect its training (or fine-tuning) data without external supervision.
This paradigm was initially adopted in robotics for segmentation of traversable terrain~\cite{dahlkamp2006self,stavens2006self,lookingbill2006reverse}, then applied to other tasks such as grasping \cite{Mar:icra:2015,Levine:ijrr:2018} and long-range sensing for navigation~\cite{ Dhiraj:iros:2017,Hadsell:jfr:2009, Nava:ral:2021}.
%

%
In summary, we present the following contributions:
\begin{itemize}
    \item a \ac{tcn} architecture to classify 3 walking modes.
    \item an approach to estimate walking modes within a time window including past, current, and future timesteps.
    \item a self-labeling procedure that allows the model to adapt to new users wearing the exosuit.
\end{itemize}

Experimental results on a multi-user dataset show that: ($i$) the TCN-based architecture outperforms previous work~\cite{zhang:tmec:2024}; ($ii$) forecasts of future values of the target variable outperform repeating the prediction for the current timestep, showing that our prediction is solid and can be useful for control purposes; ($iii$) past estimates make our self-labeling procedure effective at adapting the model's performance to a new user without any external supervision.
Finally, we integrate the classifier with the exosuit's closed-loop controller in an online experiment, validating its capability with a single subject.


\section{Problem formulation and approach}\label{sec:approach}

%
%

\subsection{Walking mode classification}
We propose an \ac{ml}-based solution that learns, from previous walking experiences, the user's past, current, and future behavior.
Our approach is a classifier estimating three walking modes that humans use when moving in indoor and urbanized environments, i.e. \ac{sa}, walking on the \ac{lg}, and \ac{sd}.
Such walking modes are represented, at the current time sample $k$, by the following discrete variable:
\begin{equation}
    c_k \in \{ \text{\ac{sa}}, \text{\ac{lg}}, \text{\ac{sd}} \}.
\end{equation}
We aim to estimate, at each timestep $k$, the walking class in a \emph{target window} ranging from $N$ timesteps before to $N$ after the current time.
More in detail, the vector
\begin{equation}
    \bm{y}_k = \left( c_{k-N}, \dots, c_k, \dots, c_{k+N} \right)^\top \in \mathbb{R}^{2N+1}
    \label{eq:output}
\end{equation}
is the target variable we want to estimate; $2N+1$ is the target window size.
Past estimates (from $k-N$ to $k-1$) are relevant to assess the perception performance and fine-tune the process with \ac{ssl} paradigms, as introduced in Sec.~\ref{sec:self_supervised}.
Past, current, and future estimates (from $k-N$ to $k+N$), instead, are useful for control, as explained in the context of our setup (Sec.~\ref{sec:exp_setup:controller}).

Our classifier is fed with the information about the user's motion provided by the minimal sensory equipment of light exosuits (e.g., thigh-mounted IMUs).
For each timestep $k$, such information is collected in the feature vector $\bm{f}_k \in \mathbb{R}^f$.
Our classifier uses as input the last $M+1$ samples organized in the input matrix $\bm{X}$:
\begin{equation}
    \bm{X}_k = \left( \bm{f}_{k-M}^\top, \bm{f}_{k-M+1}^\top, \dots, \bm{f}_{k}^\top \right)^\top \in \mathbb{R}^{f\times (M+1)}
\end{equation}
being $M+1$ the size of the so defined \emph{input window}.

Our classification model $\bm{m}$, for each timestep $k$, takes in input the matrix $\bm{X}_k$ and gives an estimate of the vector $\bm{y}_k$:
\begin{equation}
    \hat{\bm{y}}_k = \bm{m}(\bm{X}_k | {\cal D}),
    \label{eq:model}
\end{equation}
where the hat over the variable denotes the approximation of the true value due to the estimation process; ${\cal D}$ is the dataset used to train the classifier (see Sec.~\ref{sec:supervised}).
Optionally, the model $\bm{m}$ can be further refined by augmenting the training set with data acquired online and self-labeled following the procedure detailed in Sec.~\ref{sec:self_supervised}.

\begin{figure}
\centering
\includegraphics[width=0.9\linewidth]{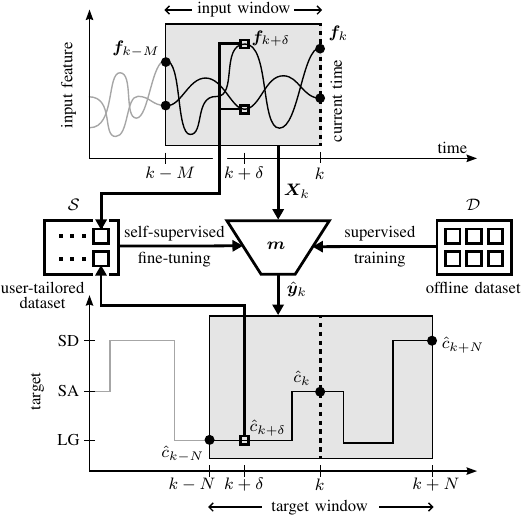}
\caption{Proposed approach: at each timestep $k$, the model $\bm{m}$ is fed with an input window of the last $M+1$ samples and outputs an estimate of walking modes in a target window including $N$ timesteps ahead and behind; $\delta$ marks the timestep in the target window where $\bm{m}$ performs best. The estimate $\hat{c}_{k+\delta}$ is used to pseudo-label the corresponding input $\bm{f}_{k+\delta}$ and refine $\bm{m}$.}
\label{fig:target_and_input_window}
\end{figure}
\subsection{Offline supervised training}\label{sec:supervised}

To train and test the classifier~\eqref{eq:model}, we organize data in sequences, each  consisting of a series of the following pairs:
\begin{equation}
    {\cal D} = \left\{ \bar{c}_{i,j}, \bm{f}_{i,j} \right\}_{i=1, j=1}^{L_j, S}
\end{equation}
where $\bar{c}$ denotes the label, i.e. the ground truth values of the walking class that are available during the data acquisition; $i$ and $j$ indicate the number of samples and sequences, respectively; $S$ is the number of data sequences, while $L_j$ is the length (expressed in timesteps) of the $j$-th sequence.

The training of the model~\eqref{eq:model} is performed offline by minimizing the distance between the model's output and the labels and using a training subset of data ${\cal D}$.
Furthermore, offline performance analysis can be carried out on a testing subset of ${\cal D}$ (different from the training set).
This analysis allows us to pick a $\delta \in [-N, N]$ such that $k+\delta$ marks the timestep in the target window where the classifier yields the best performance averaged across all samples. 
It is expected that $\delta < 0$, i.e. that the best performance is obtained within the first half of the target window, where the model can leverage input features before and after $k+\delta$, see Fig.~\ref{fig:target_and_input_window}.

\begin{algorithm}[tb!]
\caption{User-tailored self-supervised fine-tuning}\label{alg:ssl}
\begin{algorithmic}
\State $\bm{m} \gets \text{train}({\cal D})$
\State $\delta \gets \text{evaluate}(\bm{m})$
\For{a new sequence}
\State ${\cal S}=\emptyset$
\While{$k \leq H$}
\State $\bm{X}_{k} \gets \text{update}(\bm{f}_{k})$
\State $\hat{\bm{y}}_{k} = \bm{m}(\bm{X}_{k} | {\cal D})$
\State $\hat{c}_{k+\delta} \gets \text{select}(\hat{\bm{y}}_k,\delta)$ 
\State ${\cal S} \gets \text{add}(\hat{c}_{k+\delta},\bm{f}_{k+\delta})$
\EndWhile
\State $\bm{m} \gets \text{retrain}({\cal D}\cup {\cal S})$
\EndFor
\end{algorithmic}
\end{algorithm}
\subsection{User-tailored self-supervised fine-tuning}\label{sec:self_supervised}

The a-posteriori estimates provided by our classifier represent a precious source of supervision for novel sensory data acquired online.
Indeed, the estimate $\hat{c}_{k+\delta} \in \hat{\bm{y}}_{k}$ picked from the target window produced at time $k$ can be taken as the pseudo-label for the corresponding input feature vector $\bm{f}_{k+\delta}$ and create a new dataset (see Fig.~\ref{fig:target_and_input_window} for a visual reference).
Assuming that the new sequence acquired online has $H$ samples, such a new dataset ${\cal S}$ is composed of the pairs:
\begin{equation}
    {\cal S} = \{ \hat{c}_{k+\delta}, \bm{f}_{k+\delta} \}, \quad k = 1, \dots, H
\end{equation}
which can be used to augment the training set ${\cal D}$ and retrain the model $\bm{m}$.
Such a self-labeling mechanism is explained in detail in Algorithm~\ref{alg:ssl}.
First, we train a model $\bm{m}$ on the available dataset and evaluate its performance over the target window to pick a suitable $\delta$. 
Then, we consider data from a new user and for each timestep $k$, we use $\bm{m}$ to estimate the target window $\hat{\bm{y}}_k$. 
From $\hat{\bm{y}}_k$, we pick the estimate $\hat{c}_{k+\delta}$ to use as pseudo-label of the input features at the corresponding timestep, i.e. $\bm{f}_{k+\delta}$.
Once the new sequence is fully labeled, we add it to our dataset and use it to fine-tune $\bm{m}$.

\section{Experimental setup}\label{sec:exp_setup}

\subsection{Exosuit}

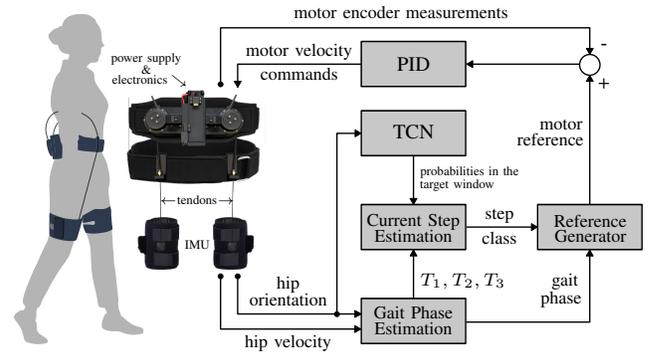
\begin{figure}
    \centering
    \resizebox{\columnwidth}{!}{%
    \footnotesize
        \begin{tikzpicture}[x=1cm,y=1cm]
            \node at (0,0) {\includegraphics[width=\columnwidth]{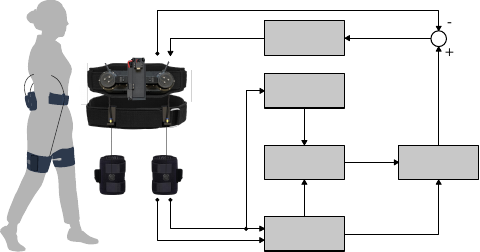}};
            \node at (1.17,1.58)[] {PID};
            \node at (1.17,0.65)[] {TCN};
            \node at (1.17,-0.58)[] {\scriptsize Current Step};
            \node at (1.17,-0.78)[] {\scriptsize Estimation};
            \node at (1.17,-1.83)[] {\scriptsize Gait Phase};
            \node at (1.17,-2.05)[] {\scriptsize Estimation};
            \node at (3.6,-0.58)[] {\scriptsize Reference};
            \node at (3.6,-0.78)[] {\scriptsize Generator};
            \node at (3.2,0.75)[] {\scriptsize motor};
            \node at (3.05,0.55)[] {\scriptsize reference};
            \node at (3.25,-1.4)[] {\scriptsize gait};
            \node at (3.20,-1.65)[] {\scriptsize phase};
            \node at (1.0,2.3)[] {\scriptsize motor encoder measurements};
            \node at (-0.4,1.75)[] {\scriptsize motor velocity};
            \node at (-0.4,1.45)[] {\scriptsize commands};
            \node at (-0.55,-1.45)[] {\scriptsize hip};
            \node at (-0.55,-1.65)[] {\scriptsize orientation};
            \node at (-0.55,-2.25)[] {\scriptsize hip velocity};
            \node at (1.85,-1.4)[] {\scriptsize $T_1, T_2, T_3$};
            \node at (1.94,0.10)[] {\tiny probabilities in the};
            \node at (1.77,-0.08)[] {\tiny target window};
            \node at (2.35,-0.5)[] {\scriptsize step};
            \node at (2.35,-0.8)[] {\scriptsize class};
            \node at (-1.8,-0.28)[] {\tiny $\leftarrow$\!\! tendons\!\! $\rightarrow$};
            \node at (-1.8,-0.9)[] {\tiny IMU};
            \node at (-2.5,1.65)[] {\tiny power supply};
            \node at (-2.5,1.5)[] {\tiny \&};
            \node at (-2.45,1.35)[] {\tiny electronics \!\!\! $\searrow$};
        \end{tikzpicture}
    }
    \caption{
    Exosuit's main components and control architecture.}%
\label{fig:exo_design}%
\end{figure}

We use a hip exosuit designed to assist hip flexion during the swing phase of walking using tendon-driven transmissions~\cite{tricomi2023environment} (Fig.~\ref{fig:exo_design}).
The device weighs \SI{2.7}{\kg} and consists of a waist belt secured at the user’s body and two textile thigh harnesses.
The belt has one actuation stage for each leg, a power supply (RED POWER battery, \SI{18.5}{\volt}, $3500$~mAh, $25$~C), and a control unit. 
Each actuation stage is driven by a compact flat brushless motor (T-Motor, AK60-6, \SI{24}{\volt}, $6$:$1$ planetary gear-head reduction, Cube Mars actuator, TMOTOR, Nanchang, Jiangxi, China) connected to a $35$~mm diameter pulley.
Assistive forces are transmitted to the user's thighs through artificial tendons made of Black Braided Kevlar Fiber (KT5703-06, maximum load capacity of $2.2$~kN, Loma Linda, CA, USA). 
The tendons run from the actuation stages at the rear of the belt to the front, where they attach distally to the thigh harnesses via 3D-printed anchor points.

The real-time control system runs on an Arduino MKR 1010 WiFi (Arduino, Ivrea, Italy), which manages gait phase estimation and actuator control at \SI{100}{\Hz}. 
The perception model is deployed on a Jetson Orin Nano (NVIDIA, Santa Clara, CA, USA), where it performs inference of the walking condition.
Hip joint kinematics are captured by two \acp{imu} (Bosch, BNO055, Gerlingen, Germany), mounted laterally on the thigh harnesses, which communicate with the control unit via Bluetooth Low Energy modules (Feather nRF52 Bluefruit, Adafruit).
%
\subsection{Closed-loop controller}
\label{sec:exp_setup:controller}
We integrate the proposed perception module with the exosuit's controller described in a previous work~\cite{Tricomi2024:nature} and skematically depicted in Fig.~\ref{fig:exo_design}. 
During the walking, a \emph{Gait Phase Estimator} uses the hip orientation and velocity provided by the IMUs to determine the start of the stance phase ($T_1$), the transition to the swing phase ($T_2$), and the end of the swing phase ($T_3$), taking into account that the swing phase is about $40\%$ of the entire gait cycle~\cite{winter2009}.
%
When $T_2$ is detected, the \emph{Current Step Estimation} block averages the class probabilities returned by the proposed walking mode classifier (implemented as a \emph{TCN}, see Sec.~\ref{sec:exp_setup:architectures}) in a time window that spans from $T_1$ (in the past) to $T_3$ (in the future).\footnote{If $T_1$ or $T_3$ fall outside the target window, we consider only the data included within it.} The class yielding the highest value is used to classify the current step (as SA, LG, or SD).
Such a step class, together with the gait phase also provided by the Gait Phase Estimation, is used by a \emph{Reference Generator} block to modulate the amplitude of the motor reference position, applying different assistance gains to match the physiological change in kinematics~\cite{riener2002}.
%
A PID control loop takes the reference motor signal to compute the commands for the exosuit's actuation system.
Note that during the stance phase (i.e., between $T_1$ and $T_2$), the motor cable is slightly released to allow the legs to extend, then it is actuated as explained above during the swing phase.

\subsection{Dataset collection}
\label{sec:exp_setup:data_collection}

The data collection campaign involved five young, healthy participants, with an average age of $25 \pm 3$ years, height of $173.9 \pm 12.0$ cm, and weight of $66.6 \pm 16.0$ kg.
Research procedures were performed according to the Declaration of Helsinki and approved by the Ethical Committee of Heidelberg University (resolution S-313/2020).

Each participant completed $6$ walking sequences, totaling $30$ sequences across all users. 
Every sequence contains instances of all three walking modes (\ac{lg}, \ac{sa}, and \ac{sd}).
In half of the sequences, we kept the exosuit off, while in the other half, we actuated it to provide constant assistance, regardless of the walking mode. 
%
%
This is done to make data more representative of a case where a perception model is used in a closed-loop fashion to control the exosuit.

For each condition (i.e., exosuit on and off): ($i$) the first sequence consists in ascending $10$ consecutive stair ramps ($10$ steps each), walking \SI{140}{\meter} on level ground, and descending the $10$ stair ramps, ($ii$) the second sequence consists in descending the $10$ stair ramps, walking \SI{40}{\meter} on level ground, and ascending the $10$ stair ramps, ($iii$) the last sequence, involves descending $8$ stair ramps, walking \SI{70}{\meter} on level ground, ascending $2$ ramps, descending $2$ ramps, walking another \SI{70}{\meter}, and finally ascending $8$ stair ramps. 

We randomized the mentioned sequences and conditions across subjects to avoid order effects.
Along the path, we recorded the left and right thighs' sagittal plane angles using IMUs.
These IMU orientations constitute the feature vector $\bm{f}_k$.
At the same time, we recorded the ground truth signal through an external manual switch used to segment the path (i.e. \ac{lg}, \ac{sa}, and \ac{sd}) by the user, which corresponds to $\bar{c}_k$.

\subsection{Classification architectures}
\label{sec:exp_setup:architectures}

\begin{figure}[t]
\centering
\includegraphics[width=\linewidth]{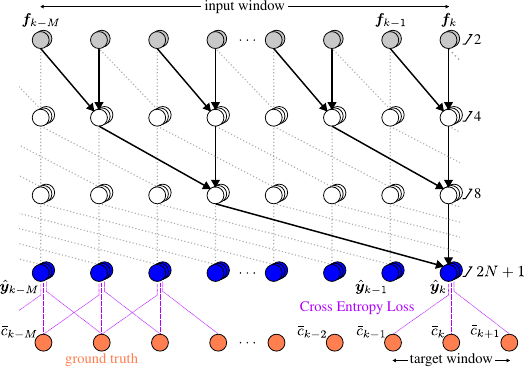}
\caption{A simplified architecture of our \ac{tcn} model with the kernel size set to $2$ and $N=1$. 
Two hidden layers output $4$ and $8$ feature channels, respectively.
The dotted gray lines represent dilated causal convolutions. Black arrows highlight those producing $\hat{\bm{y}}_{k}$ (rightmost blue circle). This is the target window estimate resulting from the two-channel input window (gray circles on top) containing $\bm{f}$ from time $k$ back to $k-M$.
}
\label{fig:tcn}
\end{figure}%

We use a \ac{tcn}~\cite{bai2018} to implement our approach described in Sec.~\ref{sec:approach}.
\ac{tcn}s are a family of architectures used to solve classification problems with sequential data, as an alternative to \ac{rnn}. 
We opt for a \ac{tcn} architecture instead of \ac{rnn}, e.g. the Long Short-Term Memory, as they perform better on many sequence-based tasks~\cite {bai2018}. 
Furthermore, their low memory footprint and computational complexity allow the deployment for real-time control even onboard low-powered devices~\cite{ingolfsson2020}.
Finally, we chose \acp{tcn} also because they can be re-trained quickly—a key aspect for our \ac{ssl} mechanism, especially when compared to other architectures such as \ac{tf}~\cite{Vaswani2017:Attention}, as detailed below.

\acp{tcn} 
perform causal convolutions, i.e., the output at a given time is convolved only with elements from earlier times in the previous layer. In this way, no information can leak from the future into the past.
These convolutions are also increasingly dilated by a factor that determines the spacing between filter elements.
The dilation factor allows for capturing long-range dependencies without increasing the network's depth.
Tuning the dilation factor and the convolution kernel sizes enables a flexible configuration of the model receptive field (i.e. the input window size defined in Sec.~\ref{sec:approach}). 

We adapt an existing Pytorch implementation~\cite{pytorch-tcn}. 
A simplification of our architecture is depicted in Fig.~\ref{fig:tcn}. 
We concatenate $3$ residual blocks with increasing feature channels to apply a dilated causal convolution. The dilation factors are set to $1$, $2$, and $4$, respectively, and the kernel size to $5$, meaning that the input window size is $57$ samples.
As the input features are collected at \SI{30}{Hz}, the result is a receptive field of about \SI{2}{\second}.
Our model's inputs are two IMU orientation signals as described in~\ref{sec:exp_setup:data_collection}.
For each input sample, the model outputs a target window of length $2N+1$, as explained in Sec.~\ref{sec:approach}.
We set $N=60$, i.e. a target window estimating \SI{2}{\second} in the past and predicting \SI{2}{\second} in the future.
Each window element contains unnormalized class scores for each of the 3 walking modes.  
Such scores are used, together with the labels, to compute the cross-entropy loss while training the model. 
At inference, the softmax function is applied to produce probability scores for each class; the highest one is used to classify the sample and produce the vector $\hat{\bm{y}}_k$.
%
%

%

%
%

For comparison purposes, we also implement a \ac{rf} classifier inspired by previous work~\cite{zhang:tmec:2024}, and a model based on the encoder–decoder \ac{tf}~\cite{Vaswani2017:Attention}.
For the \ac{rf}, although we consider only $2$ input signals (instead of $6$), we compute the same handcrafted features. In particular, we consider an input window of $60$ samples and we obtain $6$ features from each signal, respectively: first, last, minimum, maximum, average, and standard deviation values.
Note that this classifier can only produce the output at the current time (i.e., it can provide an estimate of $c_{k}$ in~\eqref{eq:output}, but not the full vector $\hat{\bm{y}}_k$) and receives a handcrafted feature set that summarizes a window of past data.
Further details of the \ac{rf} implementation 
are in Sec.~\ref{sec:results}.
The \ac{tf} model is a drop-in replacement of the \ac{tcn} model, being trained to process the same input and produce the same output. 
The \ac{tf} implementation roughly shares the same number of parameters as the \ac{tcn}, but is slower in the training process. 
In fact, the \ac{tcn} can process the entire training dataset $10$ times faster than the \ac{tf} (\SI{1}{s} per epoch on a consumer-grade GPU instead of \SI{10}{s}). 
%


\begin{figure}[!t]
\centering
\includegraphics[width=\linewidth]{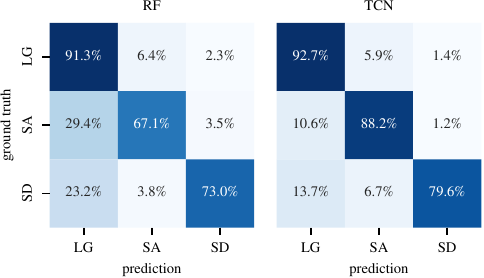}
\caption{Confusion matrix for the \ac{rf} (left) and \ac{tcn} approach (right).}
\label{fig:confusion_matrices}
\end{figure}
\subsection{Evaluation metrics}
\label{sec:approach:eval}
We use the \ac{auroc} to evaluate the performance of our classifiers. 
In particular, we use the multi-class \ac{auroc} using the one-vs-rest approach, computing the \ac{auroc} of each class against the rest and averaging the result across all classes. 
The \ac{auroc} does not depend on thresholds. It ranges between $0.5$ for a non-informative classifier and $1.0$ for an ideal one. 
We also report the confusion matrix, which summarizes predicted versus ground truth values and highlights misclassifications of the model.

%
%
%

\section{Results}\label{sec:results}

\subsection{Estimation at the current time}\label{sec:current_time}

We evaluate the performance of our \ac{tcn} approach at estimating which walking mode (i.e., \ac{lg}, \ac{sa} or \ac{sd}) is performed by a user wearing the exosuit.
The performances are compared against an \ac{rf} approach inspired by~\cite{zhang:tmec:2024}, which can only produce an estimate for the current time $k$, as explained in Sec.~\ref{sec:exp_setup:architectures}.
For this comparison, we consider only the estimate that our \ac{tcn} produces for time $k$, instead of the entire target window.
%
Also, since users may exhibit distinct walking patterns, it is valuable to analyze each individual separately, while also evaluating the model’s capacity to generalize to previously unseen users.
To this end, we apply a leave-one-user-out cross-validation procedure for both models. In each validation fold, the data of one user is excluded from training, and the \ac{auroc} is computed on that user’s data.
The averaged \ac{auroc} across all users is $0.925$ for the \ac{rf} and $0.962$ for the \ac{tcn}, confirming the superior performance and generalization capability of the proposed model. 
Figure~\ref{fig:confusion_matrices} further illustrates this by showing the aggregated confusion matrices: the \ac{tcn} notably reduces the misclassifications of the \ac{rf}, particularly for the \ac{sa} and \ac{sd} classes.
To directly compare the models on a per-user basis, we use the Wilcoxon signed-rank test. This non-parametric paired test is appropriate because it does not assume normality of the \ac{auroc} distributions and evaluates whether the differences observed for the same users are systematic. The results confirm that the \ac{tcn} yields significantly higher \ac{auroc} values than the \ac{rf} ($p = 0.03125$).

%

\begin{figure}[t]
\centering
\includegraphics[width=\linewidth]{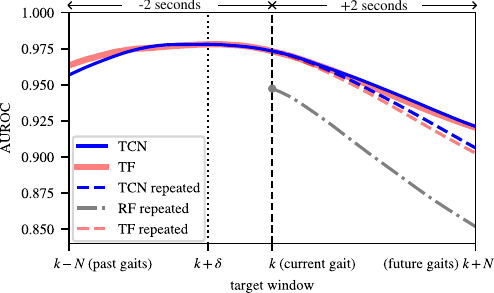}
\caption{Models' performance for each timestep in the target window: \ac{auroc} of the \ac{tcn} (solid blue) and \ac{tf} (solid red); \ac{tcn}, \ac{tf} and \ac{rf} repeating the estimate at $k$ in the future (dashed blue, red and gray dash-dotted lines, respectively). The dotted black line at timestep $k+\delta$ corresponds to the peak of performance for the \ac{tcn} model.}
\label{fig:auroc_time_window}
\end{figure}%
\begin{figure}[!t]%
\centering%
\includegraphics[width=\linewidth]{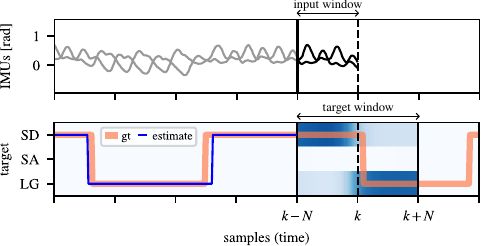}%
\caption{\ac{tcn} inference for a \SI{14}{\second} time-frame: IMUs signals composing the feature vector, with the input window (top plot, gray lines) and prediction of the model compared against the ground truth (bottom plot, blue and orange lines respectively). In the target window, the probabilities of each walking class are depicted with blue shades (the darker, the higher).}%
\label{fig:animation}%
\end{figure}%
\subsection{Estimation in a time window}\label{sec:time_window}

Our approach can provide estimation both in advance and back in time.
In practice, we use our framework to predict the future walking modes that will happen in a time window of \SI{2}{\second} and the past modes that just occurred in the last \SI{2}{\second}.
%

For this validation, we compute the \ac{auroc} averaged among all users as done in Sec.~\ref{sec:current_time} but for all the instants in the target time window, see Fig.~\ref{fig:auroc_time_window}.
The plot shows that $\delta=-10$ marks the timestep in the target window at which the classifier performs best, confirming our speculation from Sec.~\ref{sec:supervised}.
For reference, we also report the performance of the \ac{tf} based model described in Sec.~\ref{sec:exp_setup:architectures}.
As expected, all models lose performance as predictions move forward from $k$, i.e. as they forecast further in the future. However, performance values remain high (\ac{auroc} greater than $0.9$) for both models in the entire window.
Furthermore, in estimating the future, the proposed approach outperforms three trivial baselines 
built by repeating the prediction at time $k$ from $k+1$ up to $k + N$ for the \ac{tcn}, \ac{tf} and \ac{rf} models respectively. 
This confirms that the model captures hidden patterns in the data that predict the user's future behavior (that might be due to users' subtle adjustment motions before the stairs begin) and produces meaningful predictions, rather than learning to predict a constant value for future timesteps.
Finally, the plot shows that the choice between the \ac{tcn} and the \ac{tf} models is arbitrary, since their performance is similar. However, we select the \ac{tcn} for our implementation since it is more efficient to train, as explained in Sec.\ref{sec:exp_setup:architectures}.
This is relevant for our plans of fine-tuning the model onboard the exosuit device to quickly adapt to new users, leveraging our self-supervised pipeline. 

%

Figure~\ref{fig:animation} shows a time frame of the \ac{tcn} running over a sequence. The top plot displays the IMU orientations, highlighting the input window in black.
The bottom plot shows the ground truth in orange for each timestep, along with the model prediction for the current time $k$ in solid blue.
In addition, a blue colormap displays the $3$ probabilities returned by the model for each timestep in the target window. 
Darker blue corresponds to a higher probability for the corresponding class (on the $y$-axis).
The blue solid line and the target window closely match the ground truth values. 
Such a plot qualitatively confirms the classifier's good performance.

The performance of our TCN in estimating the walking mode in a time window is qualitatively shown in the video accompanying the paper.
\begin{figure}[!t]%
\centering%
\includegraphics[width=\columnwidth]{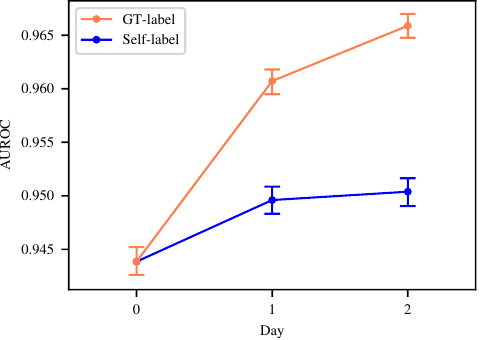}%
\caption{Models' performance over time: incrementally adding labeled data to the training set improves the \ac{auroc} (orange line) even using our self-labeling procedure (blue). Error bars denote $95\%$ confidence intervals.}%
\label{fig:SSL}%
\end{figure}%
%


%
\begin{figure*}[!t]%
\centering%
\includegraphics[width=\linewidth]{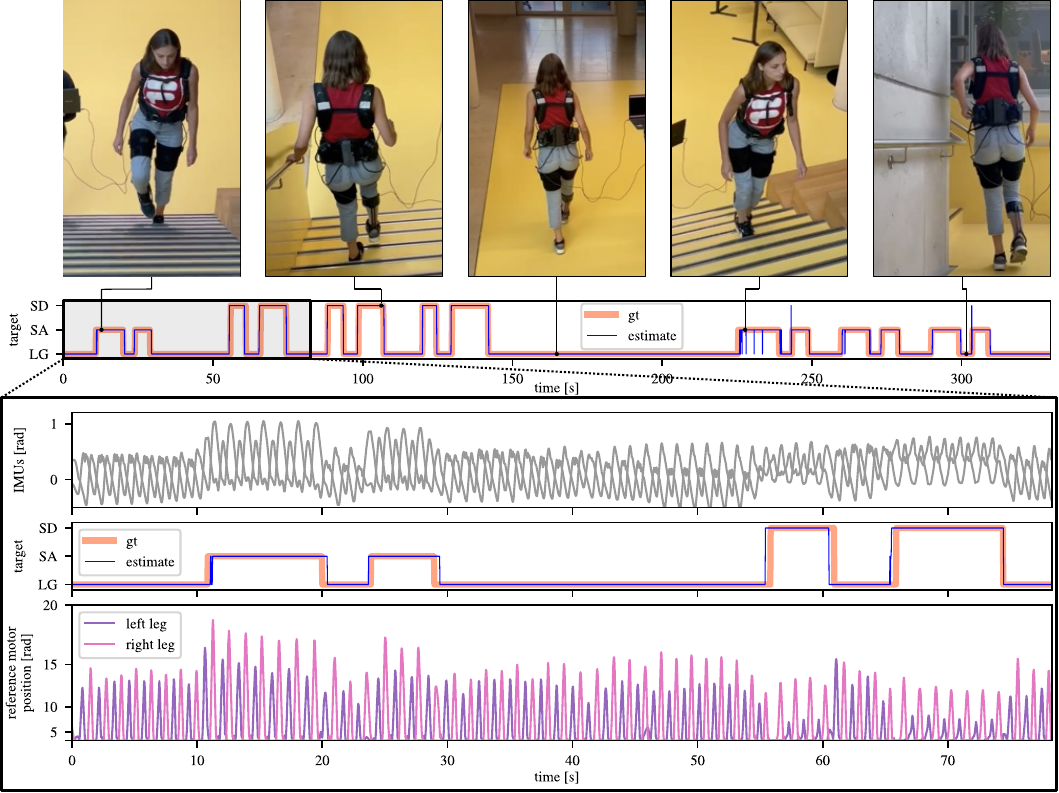}%
\caption{Snapshots of an experiment running our perception model integrated in closed-loop with the exosuit's controller. The top plot shows the recorded ground truths for the walking mode (orange) and the model's classifications (blue). The call-out rectangle at the bottom shows in detail the first part of the experiment, in particular: input features signals (top plot), walking mode ground truth and estimates (middle), and reference motor positions (bottom). The purple and pink lines show that the amplitudes of the motor references are correctly modulated according to the current walking mode.}%
\label{fig:exp4}%
\end{figure*}%
\subsection{Self-supervised learning}\label{sec:ssl}

We simulate a situation where a new user wears the exosuit over three days, and each day new walking sequences are recorded.
Additional data can be used to refine the model and improve the performance, as explained in Sec.~\ref{sec:self_supervised}.
We take the dataset recorded as described in Sec.~\ref{sec:exp_setup:data_collection}, containing walking sequences from $5$ users. Each user is kept out in turn as the \emph{new user}, then the following protocol is applied:
\begin{itemize}
    \item at day 0, we train a model using the data of $6$ sequences collected by each of the $4$ users. We compute the performance on $2$ sequences from the new user;
    \item at day 1, we augment the training set by adding the $2$ sequences for the new user from day 0. Testing is performed on $2$ new sequences from the same user.
    \item at day 2, we increase again the training set with the $2$ sequences for the new user considered in day 1. Testing is performed on $2$ new sequences from the same user.
\end{itemize}
For each user, we perform $6$ simulation runs, one for each permutation of the available sequences (i.e. $6$ per user), taken in pairs. In total, we simulate $30$ runs.
The resulting performances are averaged across all the runs for each simulated day.
%
%
Furthermore, we run the experiment twice, changing the way we label new data added to the training set each day. 
%
In the second run, we use the estimates obtained by the model trained the previous day (leveraging the self-labeling procedure of Sec.~\ref{sec:self_supervised}).
The results are shown in Fig.~\ref{fig:SSL}: the addition of novel data from a user 
improves performance on new data collected by the same user; as expected, the improvement is larger when the new data is labeled with ground truth. However, performance also improves when the proposed self-labeling procedure is adopted, which requires no external supervision and can be performed automatically.

\subsection{Closed-loop assistive robotics experiment}\label{sec:online_exp}
We run an online experiment to qualitatively assess our classifier's performance and demonstrate its practical use in closed-loop within the exosuit's controller.
%
The system was tested on a $28$-year-old female participant walking at self-selected speed along a sequence including all three modes (SA, LG, SD). 
The sequence consisted of $91$ ascending stairs, \SI{98}{m} on level ground, and $72$ descending stairs.
During the trial, IMU signals, the motor reference position command, final classifications for both legs, and ground truth were recorded.
Ground truth was obtained from a manual switch pressed by the participant when changing modes.
We deployed a model trained on the dataset described in Sec.~\ref{sec:exp_setup:data_collection}, expanded with as little as \SI{10}{\minute} of data previously collected with the new testing user.
The results are qualitatively presented in Fig.~\ref{fig:exp4}, demonstrating the performance of our approach in a real-life scenario.
The motor reference amplitudes increased during stair ascent and decreased during stair descent according to the perception module classification, which closely matches the ground truth walking mode.

\section{Conclusions and future work}\label{sec:conclusion}

We have proposed a machine learning approach to estimate the walking modes of a user wearing a lightweight exosuit.
The approach consists of a classifier implemented using a temporal convolutional network, taking the input of two IMUs and estimating the walking modes in a time window, which includes past and future time steps.
A thorough analysis has shown the performance of our approach at improving previous implementations, forecasting future walking modes,
and computing past estimates to enable our effective self-labeling procedure. 
An online proof of concept integrating the classifier in closed-loop with the exosuit's controller has shown high performance with a single subject experiment.
Overall, these results highlight the potential of user-tailored machine learning to enhance adaptability and usability in wearable assistive devices.
However, the study is not without limitations, which in turn point toward promising avenues for future research. 
The current validation involved a relatively small pool of healthy subjects, which restricts generalization. Extending the framework to a larger and more diverse population, including users with mobility impairments, is a crucial next step.
This would enable reporting quantitative measures such as inter-subject variability and prediction latency to better characterize real-time performance.
Moreover, while we demonstrated reliable control of the exosuit, we did not yet quantify whether the provided assistance improves walking performance or comfort, an aspect that will require dedicated evaluation metrics. 
The set of walking modes considered was limited to level walking and stairs, 
but our system is readily extensible to additional modes (e.g., ramps), enabling evaluation in more complex locomotion tasks.
Also, the model fine-tuning based on self-supervision has been tested only offline. Migrating the procedure fully onboard will facilitate field testing and user adaptation in real-world conditions.
Future work will investigate confidence-based pseudo-label selection and corresponding retraining strategies to improve robustness and reproducibility in online adaptation.
Finally, we plan to integrate lightweight auxiliary sensors (e.g., a barometer) for retrospective data labeling, as their long-term accuracy can improve training, even though their short-term noise limits their use for online walking mode prediction.
%

\addtolength{\textheight}{-5.5cm}  

\bibliographystyle{IEEEtran}
\bibliography{bibliography}

\end{document}